\documentclass[nohyperref]{article}

\usepackage{microtype}
\usepackage{graphicx}
\usepackage{subfigure}
\usepackage{booktabs} 

\usepackage{hyperref}
\usepackage{natbib}

\usepackage[accepted]{icml2022}

\usepackage{amsmath}
\usepackage{amssymb}
\usepackage{mathtools}
\usepackage{amsthm}

\usepackage[capitalize,noabbrev]{cleveref}

\theoremstyle{plain}

\theoremstyle{definition}

\theoremstyle{remark}

\usepackage{multirow}
\usepackage{threeparttable}

\usepackage[textsize=tiny]{todonotes}

\icmltitlerunning{Density-Aware Hyper-Graph Neural Networks 
for Graph-based Semi-supervised Node  Classification}

\begin{document}

\twocolumn[
\icmltitle{Density-Aware Hyper-Graph Neural Networks 

for Graph-based Semi-supervised Node Classification}



\begin{icmlauthorlist}
\icmlauthor{Jianpeng Liao}{sch}
\icmlauthor{Qian Tao}{sch}
\icmlauthor{Jun Yan}{sch2}
\end{icmlauthorlist}

\icmlaffiliation{sch}{School of Software, South China University of Technology, Guangzhou, China}
\icmlaffiliation{sch2}{Concordia Institute for Information Systems Engineering, Concordia University, Montreal, Quebec, Canada}

\icmlcorrespondingauthor{Jianpeng Liao}{sejianpengliao@mail.scut.edu.cn}
\icmlcorrespondingauthor{Qian Tao}{taoqian@scut.edu.cn}

\icmlkeywords{Semi-supervised Node Classification, hyper-graph, Graph, Graph Neural Networks}

\vskip 0.3in
]



\printAffiliationsAndNotice{}  

\begin{abstract}
Graph-based semi-supervised learning, which can exploit the connectivity relationship between labeled and unlabeled data, has been shown to outperform the state-of-the-art in many artificial intelligence applications. One of the most challenging problems for graph-based semi-supervised node classification is how to use the implicit information among various data to improve the performance of classifying. Traditional studies on graph-based semi-supervised learning have focused on the pairwise connections among data. However, the data correlation in real applications could be beyond pairwise and more complicated. The density information has been demonstrated to be an important clue, but it is rarely explored in depth among existing graph-based semi-supervised node classification methods. To develop a flexible and effective model for graph-based semi-supervised node classification, we propose a novel Density-Aware Hyper-Graph Neural Networks (DA-HGNN). In our proposed approach, hyper-graph is provided to explore the high-order semantic correlation among data, and a density-aware hyper-graph attention network is presented to explore the high-order connection relationship. Extensive experiments are conducted in various benchmark datasets, and the results demonstrate the effectiveness of the proposed approach.
\end{abstract}

\section{Introduction}
Over the last few years, Graph Neural Networks (GNNs) have attracted much attention because of its ability to effectively deal with graph-structured data and achieve amazing performance, and have been  widely used for many machine learning tasks including computer vision \cite{DBLP:conf/cvpr/LiLCCYY20, DBLP:conf/cvpr/ChenWWG19}, recommender systems \cite{DBLP:conf/icml/YuQ20, DBLP:conf/kdd/HuangDDYFW021}, neural machine translation \cite{DBLP:conf/acl/YinMSZYZL20, DBLP:conf/aaai/XiaHLS19}, and others. Compared with the traditional neural networks that encode each single data separately, the GNNs are able to encode the graph structure of different input data, which allows it to obtain more information than the single data encoding for neural network learning. Graph Convolutional Networks (GCNs) is one of the most typical GNNs, proposed by Kipf et al. \cite{DBLP:conf/iclr/KipfW17}, and provided a novel approach to graph convolutional operator design. While it can not distinguish the importance of each neighboring nodes during feature aggregation. Velickovic et al. \cite{DBLP:conf/iclr/VelickovicCCRLB18} introduced a self-attention layer into GNNs and proposed Graph Attention Networks (GATs) to specify different attention weights to different nodes during neighborhood aggregation. More recently, many improved variants of GNNs have been proposed, such as Graph Learning-Convolutional Networks (GLCN) \cite{DBLP:conf/cvpr/JiangZLTL19}, Factorizable Graph Convolutional Networks (FactorGCN) \cite{DBLP:conf/nips/YangFSW20}, Frequency Adaptation Graph Convolutional Networks (FAGCN) \cite{DBLP:conf/aaai/BoWSS21}, and so on. 

In recent years, semi-supervised learning has earned more attention because it can effectively improve the generalization ability of deep neural networks. The aim of semi-supervised learning is to use both labeled and unlabeled data to improve model performance. One of the main challenges of semi-supervised learning is how to exploit the correlation between labeled and unlabeled samples as well as the implicit information among data to improve learning performance. Graph-based semi-supervised learning methods have been shown to be one of the most effective approaches for semi-supervised node classification tasks, as they are capable of exploiting the connectivity relationship between small amounts of labeled samples and a relatively large number of unlabeled samples to improve the performance of classifying. Among existing graph-based methods, such as graph convolutional networks and label propagation, have been demonstrated as one of the most effective approaches for semi-supervised node classification \cite{DBLP:conf/cvpr/LinGL20}. 

Until now, a great many graph-based semi-supervised node classification methods have been proposed. For example, Graph Convolutional Networks (GCNs) \cite{DBLP:conf/iclr/KipfW17} performed label prediction based on graph neighborhood aggregation. Graph Learning-Convolutional Networks (GLCN) \cite{DBLP:conf/cvpr/JiangZLTL19} integrated graph learning and graph convolutional networks to learn an optimal graph representation. LP-DSSL \cite{DBLP:conf/cvpr/IscenTAC19} employed a graph label propagation network to generate pseudo-labels. Deep Graph Learning (DGL) \cite{DBLP:journals/pr/LinKLZC21} used deep graph learning networks to explore global and local graph structures. However, these methods all only focused on the pairwise connections among data on the graph structure. The data correlation in real practice could be beyond pairwise relationship and even far more complicated. Under such circumstances, only exploring the pairwise connections and modeling it as a graph may lose the high-order semantic correlation among data, especially for complicated data such as image datasets and citation network datasets. Traditional graph structure cannot fully formulate the data correlation and thus limits the application of GNNs models \cite{DBLP:conf/aaai/FengYZJG19}. 

Hyper-graph is a generalization of graph, in which each hyperedge can link to any number of nodes, and therefore, compared with the simple graph, hyper-graph can more effectively represent the high-order semantic relationship among data. Similar to GCNs \cite{DBLP:conf/iclr/KipfW17}, Feng et al. \cite{DBLP:conf/aaai/FengYZJG19} proposed Hyper-Graph Neural Networks (HGNN) to perform hyper-graph convolution. Inspired by GATs \cite{DBLP:conf/iclr/VelickovicCCRLB18}, Chen et al. \cite{DBLP:conf/trustcom/ChenCLW20} proposed Hyper-Graph Attention Networks (HGATs) to aggregate neighborhood features with different attention weights. So far, many variant models of hyper-graph have been proposed, including Dynamic Hyper-Graph Neural Networks (DHGNN) \cite{DBLP:conf/ijcai/JiangWFCG19}, Hyper-Graph Label Propagation Networks (HLPN) \cite{DBLP:conf/aaai/ZhangWCZWZ020}, and others. In our work, we introduce hyper-graph to explore the high-order semantic correlation among data and perform semi-supervised node classification based on hyper-graph representation learning.

One of the most challenging problems for semi-supervised learning is how to exploit the implicit information among data to improve model performance. However, existing graph-based semi-supervised node classification methods most only utilized the obvious graph structure information. Recently, some implicit information among data, for example, density information, has been demonstrated to provide important clues for semi-supervised node classification \cite{DBLP:conf/cvpr/LiLCCYY20}, yet it is rarely exploited in depth. Meanwhile, among most of the existing graph-based semi-supervised node classification methods, the model learning performance degrades quickly with a diminishing number of labeled samples per class. When the labeled samples are extremely limited, the model performance is unsatisfactory. Li et al. \cite{DBLP:conf/cvpr/LiLCCYY20} firstly exploited the density information explicitly for graph-based deep semi-supervised visual recognition. Yet it is also only based on the exploration of graph-structure relationship among data, while high-order correlation has been ignored. Inspired by Li et al. \cite{DBLP:conf/cvpr/LiLCCYY20}, we decide to explore density information among data on hyper-graph structure to improve semi-supervised node classification accuracy in this work.

In this paper, we propose a novel Density-Aware Hyper-Graph Neural Networks (DA-HGNN) for graph-based semi-supervised node classification. In our approach, hyper-graph is provided to explore the high-order semantic correlation among data. Meanwhile, we exploited density information among data explicitly to improve semi-supervised node classification performance. We defined a density rule for each node and hyperedge. 
We learn the density information and integrate it as a part of attention weight in each layer. Intuitively, the higher the density of a node or a hyperedge, the more attention we need to pay to it when performing neighborhood aggregation. By fusing the density information, we can exploit it explicitly for hyper-graph representation learning. 
More specifically, the proposed DA-HGNN first adopts a single-layer hyper-graph convolutional network to generate low-dimensional node and hyperedge feature embeddings, and then employs a density-aware hyper-graph attention network to explore the high-order relationship among data. The density-aware hyper-graph attention layer consists of two parts, density-aware attention vertex aggregation and density-aware attention hyperedge aggregation, which first gather node features to hyperedges to enhance hyperedge embedding, and then aggregate the connected hyperedge features to generate a new vertex embedding. Through this node-hyperedge-node feature transform mechanism, we can efficiently explore the high-order semantic correlation among data. The main contributions of our work can be summarized as follow:

\begin{itemize}
\item We propose a novel Density-Aware Hyper-Graph Neural Networks (DA-HGNN) to exploit density information explicitly on hyper-graph structure for graph-based semi-supervised node classification. We define a density rule for each node and hyperedge, and then integrate the density information as a part of attention weight. Based on the density-aware attention weight, we can perform more accurate hyper-graph neighborhood aggregation.
\item We have conducted extensive experiments on three image datasets and demonstrated the validness of density-aware attention weight and the effectiveness of the proposed DA-HGNN on semi-supervised node classification tasks.
\end{itemize}

\section{Related Work}

\textbf{Graph Neural Networks.}  
The Graph Neural Networks (GNNs) is one of neural networks designed for processing graph-structured data. The core idea of GNNs is graph information propagation \cite{DBLP:journals/tnn/WuPCLZY21}, which can be divided into spectral-based approaches \cite{DBLP:conf/nips/DefferrardBV16,DBLP:conf/iclr/KipfW17,DBLP:journals/tsp/LevieMBB19} and spatial-based approaches \cite{DBLP:conf/icml/NiepertAK16,DBLP:conf/icml/GilmerSRVD17}. 
Graph Convolutional Networks (GCNs) \cite{DBLP:conf/iclr/KipfW17} is one of the most classic GNNs, which provided a novel idea for graph spectral information propagation and played a central role in building up many other complex GNNs models.  
Transductive Propagation Networks (TPN) \cite{DBLP:conf/iclr/LiuLPKYHY19} performed meta-learning transductive labels propagation on graph.
Jiang and Lin et al. \cite{DBLP:conf/cvpr/JiangZLTL19,DBLP:journals/pr/LinKLZC21} suggested using a weight vector to fit the similarity between nodes, and integrated the graph learning and the downstream GNNs to learning an optimal graph structure.
Factorizable Graph Convolutional Networks (FactorGCN) \cite{DBLP:conf/nips/YangFSW20} tended to disentangle the heterogeneous intertwined relations encoded in a graph. 
Latterly, by introducing a self-gate mechanism, Frequency Adaptation Graph Convolutional Networks (FAGCN) \cite{DBLP:conf/aaai/BoWSS21} could adaptively integrate different signals in the process of message propagation. 
Gallicchio and Li et al. \cite{DBLP:conf/aaai/GallicchioM20, DBLP:conf/icml/Li0GK21, DBLP:conf/iccv/Li0TG19} all advised to utilize deep graph neural networks to explore deep graph information propagation.

\textbf{Hyper-Graph Neural Networks.} Hyper-graph is a generalization of graph. Compared with the simple graph in which an edge can only connect two vertices, each hyperedge in a hyper-graph can link any number of nodes, which makes it can explore the high-order relationship among data. Hyper-Graph induced Convolutional Networks \cite{DBLP:journals/tnn/ShiZZMZGS19} adopted a hyper-graph learning process to optimize the high-order correlation among visual data. Afterwards, Hyper-Graph induced Convolutional Manifold Networks (H-CMN) \cite{DBLP:conf/ijcai/JinCZSDJ19} improved the representation capacity of deep convolutional neural networks for the complex data by combining with mini-batch-based manifold preserving method. Hyper-Graph Neural Networks (HGNN) \cite{DBLP:conf/aaai/FengYZJG19} performed node-edge-node transform through hyperedge convolution operations to better refine the features representation. Recently, many improved methods have been proposed, including Dynamic Hyper-Graph Neural Networks (DHGNN) \cite{DBLP:conf/ijcai/JiangWFCG19}, Hyper-Graph Attention Networks (HGATs) \cite{DBLP:conf/trustcom/ChenCLW20}, Hyper-Graph Label Propagation Networks (HLPN) \cite{DBLP:conf/aaai/ZhangWCZWZ020}, Hyper-Graph Enhanced Graph Reasoning \cite{DBLP:conf/icml/ZhengYG021}, and so on.

\textbf{Attention in Graph Neural Networks.} Although general graph neural networks can perform feature aggregation through graph information propagation, it cannot distinguish the importance of each neighboring node. Attention mechanism is an effective technique allowing the neural networks to select key features in the data  \cite{DBLP:conf/icml/BertasiusWT21,DBLP:conf/icml/JaegleGBVZC21}. 
Lately, attention mechanism was led into GNNs models and gradually became state-of-the-art. 
By adopting a self-attention layer, Graph Attention Networks (GATs) \cite{DBLP:conf/iclr/VelickovicCCRLB18} enabled specifying different attention weights to different nodes in a neighborhood and performed more accurate feature aggregation. 
Inspired by GATs \cite{DBLP:conf/iclr/VelickovicCCRLB18}, Hyper-Graph Attention Networks (HGATs) \cite{DBLP:conf/trustcom/ChenCLW20} introduced the attention mechanism into the hyper-graph neural networks to encode the high-order data relation. 
Shi et al. \cite{DBLP:conf/ijcai/ShiHFZWS21} suggested combining the multi-head attention into label propagation algorithm. 
Dasoulas et al. \cite{DBLP:conf/icml/DasoulasSV21} introduced LipschitzNorm to enforce the attention model to be Lipschitz continuous, which can solve the gradient explosion problem of deep graph attention networks. 
In our work, inspired by Velickovic, Chen and Li et al. \cite{DBLP:conf/iclr/VelickovicCCRLB18, DBLP:conf/trustcom/ChenCLW20, DBLP:conf/cvpr/LiLCCYY20}, we suggest the integration of density information, attention mechanism and hyper-graph neural networks to achieve more effective representation learning.

\textbf{Graph-based Semi-supervised Node Classification.} Graph-based semi-supervised learning methods have been shown to be one of the most effective approaches for semi-supervised node classification \cite{DBLP:conf/cvpr/LinGL20}, as they are capable of exploiting the connectivity relationship between small amounts of labeled samples and a relatively large number of unlabeled samples to improve classification performance.
Graph-based methods, such as graph convolutional networks and label propagation, have been demonstrated as one of the most effective approaches for semi-supervised node classification \cite{DBLP:conf/cvpr/LinGL20}. In recent, many graph-based variant models have been proposed for semi-supervised node classification tasks. 
Graph Learning-Convolutional Networks (GLCN) \cite{DBLP:conf/cvpr/JiangZLTL19} introduced a graph learning module to learn an optimal graph structure which makes GCNs \cite{DBLP:conf/iclr/KipfW17} better for semi-supervised learning. Li et al. \cite{DBLP:conf/cvpr/LiW00G19} suggested proper low-pass graph convolutional filters to generate smooth and representative nodes features, and took data features as signals to perform label propagation operations simultaneously. Iscen et al. \cite{DBLP:conf/cvpr/IscenTAC19} proposed a graph-based transductive label propagation to generate pseudo-labels and combine it with deep inductive framework. Deep Graph Learning (DGL) \cite{DBLP:journals/pr/LinKLZC21} suggested exploring global and local graph structure to improve semi-supervised node classification performance. Li et al. \cite{DBLP:conf/cvpr/LiLCCYY20} suggested using the density information explicitly, which is effective for deep semi-supervised node classification. 
 
In this paper, we introduce hyper-graph to explore the high-order semantic correlation among data, then propose DA-HGNN to exploit density information and incorporate it for hyper-graph representation learning.

\section{Methods}

\textbf{Overview.} 
In this work, we exploit density information explicitly by integrating it into hyper-graph neural networks, and propose a novel Density-Aware Hyper-Graph Neural Networks (DA-HGNN). The proposed DA-HGNN first adopts a single-layer hyper-graph convolutional network to generate a low-dimensional node and hyperedge feature embedding, and then employs a density-aware hyper-graph attention network to explore the high-order connectivity relationship among data. 
We define a density rule for each node and hyperedge, then compute the density information and integrate it as a part of attention weight in each density-aware hyper-graph attention layer and perform neighborhood aggregation. 
The density-aware hyper-graph attention layer consists of two parts, density-aware attention vertex aggregation and density-aware attention hyperedge aggregation, which first gathers node features to hyperedges to enhance hyperedge features and then aggregates the connected hyperedge features to generate new vertex features. Through this node-hyperedge-node feature transform mechanism, we can efficiently explore the high-order semantic correlation among data. More detail will be explained as follows.

\subsection{Hyper-Graph Construction}

A hyper-graph can be formulated as $\mathcal{G}=(\mathcal{V}, \mathcal{E})$, which includes a vertex set $\mathcal{V}$ and a hyperedge set $\mathcal{E}$. Let $X=(x_{1},x_{2},\dots,x_{n})\in \mathbb{R}^{n \times d} $ be the collection of $n$ data vectors of $d$ dimension, where $x_{i}$ denotes the feature vector of $i$-th sample. The structure of hyper-graph can be denoted by an incidence matrix $H\in \mathbb{R}^{n \times m}$, in which $H(x_{i},e_{k})=1$ indicates that the node $x_{i}$ is connected by the hyperedge $e_{k}$, otherwise $H(x_{i},e_{k})=0$, and $n$ and $m$ are the numbers of nodes and hyperedges, respectively, in the hyper-graph. 

In this paper, we construct a hyper-graph following the previous work HGNN \cite{DBLP:conf/aaai/FengYZJG19} and HGATs \cite{DBLP:conf/trustcom/ChenCLW20}. 
The similarity between samples is the key basis for hyper-graph construction. We adopt a Euclidean distance to measure the similarity between nodes. 
We first compute the Euclidean distance between every two samples and obtain the Euclidean distance matrix $D \in \mathbb{R}^{n \times n}$, where each element $d(x_{i},x_{j})$ denotes the Euclidean distance between node  $x_{i}$ and  $x_{j}$. 
Then, for each node, we select its top-$k$ nearest neighbors to construct a hyperedge that connects to $k+1$ nodes. 
Finally, we obtain a hyper-graph incidence matrix $H\in \mathbb{R}^{n \times m}$, where each element $H(x_{i},e_{k})$ represents the connection relationship of node  $x_{i}$ and  $e_{k}$. $n$ and $m$ are the numbers of nodes and hyperedges respectively, and $n=m$ in implementation.

\subsection{Density-Aware Hyper-Graph Neural Networks}

The proposed DA-HGNN incorporates the density information into hyper-graph neural networks and integrates it as a part of attention weights, and then performed density-aware attention neighborhood aggregation. It mainly consists of a single-layer hyper-graph convolutional network and a two-layer density-aware hyper-graph attention network.

The single-layer hyper-graph convolutional network is employed to learn a low-dimensional embedding of node feature and hyperedge feature. This is because when the dimension $d$ of the node feature matrix $X$ is large, the density-aware hyper-graph attention network will have high computational overhead and may be less effective due to the long attention weight vector to be trained. 
By inserting a hyper-graph convolutional layer before the density-aware hyper-graph attention network can effectively solve this problem.
The inputs of the hyper-graph convolutional network include the initial node feature matrix $X_{0}\in \mathbb{R}^{n \times d_{0}}$ and the hyper-graph incidence matrix $H\in \mathbb{R}^{n \times m}$. Through the hyper-graph convolution operation, the hyper-graph convolutional network can perform node-hyperedge-node feature transform, and then the low-dimensional node feature embedding  $X\in \mathbb{R}^{n \times d}$ and hyperedge feature embedding  $E\in \mathbb{R}^{m \times d}$ can be obtained. It can be formulated as follows: 
\begin{equation}
     X = D_{e}^{-1/2} H^{\top } D_{v}^{-1/2} X_{0} \Theta
\end{equation}
\begin{equation}
     E = D_{v}^{-1/2} H D_{e}^{-1/2} X
\end{equation}

where $D_{e}$ and $D_{v}$ denote the diagonal matrices of the hyperedge degrees and the vertex degrees, respectively. $\Theta$ is a trainable weight matrix. $\cdot ^{\top }$ denotes transposition.

The inputs of the density-aware hyper-graph attention network inlcude the low-dimensional node feature embedding  $X\in \mathbb{R}^{n \times d}$, hyperedge feature embedding  $E\in \mathbb{R}^{m \times d}$ otained from the hyper-graph convolutional network and the hyper-graph incidence matrix $H\in \mathbb{R}^{n \times m}$. And then performed hyper-graph attention neighborhood aggregation. The density-aware hyper-graph attention network mainly consists of two parts, density-aware attention vertex aggregation and density-aware attention hyperedge aggregation. The density-aware attention vertex aggregation module is employed to aggregate the information of connected vertex to the hyperedge. Similarly, the density-aware attention hyperedge aggregation module is applied to aggregate the hyperedge information to enhance the node representation. More detail about this will be described as follows.

\textbf{Density-aware attention vertex aggregation.}
Density-aware attention vertex aggregation module integrates the density information of each node as a part of attention weights, and then performs attention vertex aggregation. Firstly, we define a density for each node, which can be defined as the sum of the similarities of neighbor nodes whose similarity with the target node is greater than a predefined threshold. The density of node $x_{i}$ can be formulated as
\begin{equation}
    \rho_{x_{i}}=\sum_{x_{k} \in \mathcal{N}\left(x_{i}\right)}
    \left\{\begin{array}{l}
    \begin{matrix}
    \operatorname{sim}\left(x_{i}, x_{k}\right), & \text{ if } \operatorname{sim}\left(x_{i}, x_{k}\right) > \delta \\
    0, & \text{ if } 
    \operatorname{sim}\left(x_{i}, x_{k}\right) \le \delta
\end{matrix}
\end{array}\right.
\end{equation}
where $\mathcal{N}\left(x_{i}\right)$ denotes the neighbors set of node $x_{i}$. $\delta$ is a predefined threshold. $\operatorname{sim}(\cdot)$ is a similarity measure function, which can adopt Cosine similarity in implementation. 

Intuitively, the higher the density of a node, the more neighbors that are more similar to it. In other words, the target node is in a more densely distributed area. Based on the density-peak assumption \cite{rodriguez2014clustering}, the nodes with higher density are closer to the cluster center. Therefore, higher weights need to be assigned when performing neighborhood feature aggregation. By combing the density information with the traditional attention mechanism based on feature similarity to obtain the final attention weights, we are able to achieve more accurate attention neighborhood aggregation.

In the density-aware attention vertex aggregation module, we compute the attention weight of each node relative to the hyperedge it is on. We employ an attention mechanism with the similar characteristics of GATs \cite{DBLP:conf/iclr/VelickovicCCRLB18}. 
Firstly, we design a learnable linear transformation weight matrix $W$ to project the node and hyperedge features into high-level, which can be implemented by multiplying matrix $W$. 
Afterwards, we adopt an attention mechanism $\operatorname{Attention}(\cdot)$ to calculate the attention value between node $x_{i}$ and hyperedge $e_{k}$, which can represent as
\begin{equation}
    a_{x_{i},e_{k}} = \operatorname{Attention}(Wx_{i}, We_{k})
\end{equation}
Then we combine the density information with the attention value of node $x_{i}$ to obtain the final attention weight $\tilde{a}_{x_{i},e_{k}}$, which can formulated as
\begin{equation}
    \tilde{a}_{x_{i},e_{k}} = a_{x_{i}, e_{k}}+\tilde{\rho}_{x_{i}}
\end{equation}
where $\tilde{\rho}_{x_{i}} \in [0,\operatorname{max}(a_{X})]$ is the normalized density, and $a_{X}$ is the collection of attention value $a_{x_{i},e_{k}}$. 
Finally, we perform a probabilistic transformation on each attention weight to obtain the final attention coefficient $\text{coe}_{x_{i},e_{k}}$, whihc can formulated as
\begin{equation}
    \text{coe}_{x_{i},e_{k}}=\frac{\exp \left(\tilde{a}_{x_{i},e_{k}} \right)}{\sum_{x_{j} \in \mathcal{N}\left(e_{k}\right)} \exp \left(\tilde{a}_{x_{i},e_{k}} \right)}
\end{equation}
where $\mathcal{N}\left(e_{k}\right)$ denotes the set of vertices connected by the hyperedge $e_{k}$. The attention coefficient means the importance of node $x_{i}$ in the process of neighborhood feature aggregation to generate hyperedge embedding $e_{k}$. The adoted attention mechanism $\operatorname{Attention}(\cdot )$ can be designed similar to the GATs \cite{DBLP:conf/iclr/VelickovicCCRLB18}. We first concatenate the node embedding vector and the hyperedge embedding vector, and then employ a weight vector $\alpha_{X} \in \mathbb{R}^{2d \times 1}$ to map it to a scalar value. It can be formulated as
\begin{equation}
\begin{array}{l}
    \text{coe}_{x_{i},e_{k}}= \\
    \quad \frac{\exp \left(\operatorname{LeakyReLU}\left(\alpha_{X} ^{\top }\left(W x_{i} \parallel  W e_{k}\right)\right)+\tilde{\rho}_{x_{i}}\right)}
    {\sum_{x_{j} \in \mathcal{N}\left(e_{k}\right)} \exp \left(\operatorname{LeakyReLU}\left(\alpha_{X}^{\top}\left(W x_{j} \parallel  W e_{k}\right)\right)+\tilde{\rho}_{x_{j}}\right)}
\end{array}
\end{equation}
where $\operatorname{LeakyReLU}(\cdot)$ is an activation function to apply a nonlinearity. And $\parallel$ represents the concatenation operation. Then we can obtain the attention coefficient matrix $\mathit{COE}_{X}\in \mathbb{R}^{n \times m}$, of which each element is $\text{coe}_{x_{i},e_{k}}\in \left [0,1 \right ] $, where $n$ and $m$ are the number of nodes and hyperedges,  respectively. At last, we utilize this attention coefficient matrix to perform feature aggregation, which is formulated as
\begin{equation}
    \tilde{E}  = \sigma (\mathit{COE}_{X}^{\top }WX)
\end{equation}
where $\sigma(\cdot)$ is an activation function, which can be $\operatorname{ELU}(\cdot)$ in implementation, and $\tilde{E}\in \mathbb{R}^{m \times d^{'}}$.

\textbf{Density-aware attention hyperedge aggregation.} 
Density-aware attention hyperedge aggregation module integrates the density information of each hyperedge as a part of the attention weight, and then aggregates the connected hyperedge to enhance the node embedding. Similar to the density-aware attention vertex aggregation, firstly, we need to define a density for each hyperedge. The density of each hyperedge can be defined as the sum of the density of all nodes connected by this hyperedge, which can be formulated as
\begin{equation}
    \rho_{e_{k}}=\sum_{x_{j} \in \mathcal{N}\left(e_{k}\right)} \rho_{x_{j}}
\end{equation}
where $\mathcal{N}\left(e_{k}\right)$ denotes the set of vertices connected by the hyperedge $e_{k}$.

Intuitively, the higher the density of a hyperedge, which means this hyperedge is distributed in node-dense area. Thus, when perform hyperedge feature aggregation, we need to pay more attention to it. While traditional attention mechanisms only consider feature similarity, which may be sub-optimal. By fusing the density information, it can effectively avoid this defect and aggregate hyperedge features more accurately.

In the density-aware attention hyperedge aggregation module, we compute the attention weight of each hyperedge relative to each node connected by this hyperedge. We employed an attention mechanism similar to the density-aware attention vertex aggregation module, which can be formulated as follows: 
\begin{equation}
\begin{array}{l}
\text{coe}_{e_{k}, x_{i}}= \\
\quad \frac{\exp \left(\operatorname{LeakyReLU}\left(\alpha_{E}^{\top }\left(W e_{k} \parallel W x_{i} \right)\right)+\tilde{\rho}_{e_{k}}\right)}
{\sum_{e_{j} \in \mathcal{N}\left(x_{i}\right)} \exp \left(\operatorname{LeakyReLU}\left(\alpha_{E}^{\top}\left(W e_{j} \parallel W x_{i} \right)\right)+\tilde{\rho}_{e_{\mathbf{j} }}\right)}
\end{array}
\end{equation}
where $\mathcal{N}\left(x_{i}\right)$ represents the set of hyperedges connecting to vertex $x_{i}$. $\alpha_{E} \in \mathbb{R}^{2d \times 1}$ is a weight vector to be trained. And $\tilde{\rho}_{e_{\mathbf{k} }}$ is the normalized density.

Afterwards, we can obtain the attention coefficient matrix $\mathit{COE}_{E}\in \mathbb{R}^{m \times n}$, of which each element is $\text{coe}_{e_{k},x_{i}}\in [0,1] $. Finally, we utilize this attention coefficient matrix to perform hyperedge feature aggregation, which can be formulated as
\begin{equation}
    \tilde{X}  = \sigma (\mathit{COE}_{E}^{\top } \tilde{E})
\end{equation}
where $\sigma(\cdot)$ is an activation function, which can be implemented as $\operatorname{ELU}(\cdot)$, and $\tilde{X}\in \mathbb{R}^{n \times d^{'}}$.

We combine the two modules described above to form a density-aware hyper-graph attention layer. In each density-aware hyper-graph attention layer, we first pay a density-aware attention weight to each node and gather node features to enhance hyperedge features, and then we assign a density-aware attention weight to each hyperegde and aggregate the connected hyperedge features to generate new vertex features. The density-aware hyper-graph attention layer can be formulated as
\begin{equation}
\tilde{X}=\operatorname{ELU}\left(\mathit{COE}_{E}^{\top } \ \operatorname{ELU}\left(\mathit{COE}_{X}^{\top } W X \right)\right)
\end{equation}
where $\operatorname{ELU}(\cdot)$ is an activation function. More specific algorithm of the density-aware hyper-graph attention layer is summarized in \cref{DA-HGAT}.
By using this node-hyperedge-node feature transform mechanism, we can efficiently explore the high-order semantic correlation among data.

The proposed DA-HGNN first adopts a single-layer hyper-graph convolutional network to generate low-dimensional node and hyperedge feature embeddings, and then employs a two-layer density-aware hyper-graph attention network to explore the high-order connection relationship. In implementation, we apply a multi-head attention for the first density-aware hyper-graph attention layer to enhance the feature aggregation. The output feature representation of this layer is obtained by concatenating the output features of each head, which can be formulated as follow:
\begin{equation}
\tilde{X}= \parallel_{s=1}^{S} \operatorname{ELU}\big(\mathit{COE}_{E}^{\top } \ \operatorname{ELU}(\mathit{COE}_{X}^{\top } W X)\big)
\end{equation}
where $\parallel_{s=1}^{S}$ denotes the concatenation operation, and $S$ is the number of attention heads. For the second density-aware hyper-graph attention layer, we adopted a one-head atteneion. The final output of the DA-HGNN is a low-dimensional node feature embedding, and the class prediction $Z \in \mathbb{R}^{n \times c}$ can be obtained by performing a $\operatorname{softmax}(\cdot)$, where $n$ and $c$ are the number of samples and classes, respectively.

We adopt a cross-entropy loss as the optimization function, which can be formulated as
\begin{equation}
\mathcal{L}=-\sum_{i \in L} \sum_{j=1}^{c} Y_{i j} \ln Z_{i j}
\end{equation}
where $Y$ is the truth label matrix, and $L$ is the set of labeled samples.

\SetKwComment{Comment}{//}{}
\begin{algorithm}
  \caption{Density-aware hyper-graph attention layer.}
  \label{DA-HGAT}
    \KwIn{Node feature embedding  $X\in \mathbb{R}^{n \times d}$, hyperedge feature embedding  $E\in \mathbb{R}^{m \times d}$, hyper-graph incidence matrix $H\in \mathbb{R}^{n \times m}$.} 
    
    \KwSty{Initialize} $W$, $\alpha_{X}$, $\alpha_{E}$\\
    $\mathit{WX} = X\times W$\\
    $\mathit{WE} = E\times W$\\
    
    $\rho_{X}, \rho_{E} = f_{d}(\mathit{WX},H)$    
    \begin{scriptsize}\Comment*[f]{compute density}\end{scriptsize}
    
    $a_{X} = f_{a_X}(\mathit{WX},\mathit{WE},\alpha_{X})$
    \begin{scriptsize}\Comment*[f]{compute attention}\end{scriptsize}\\
    $\tilde{\rho}_{X} = f_{n}(\rho_{X},a_{X})$
     \begin{scriptsize}\Comment*[f]{normalize density}\end{scriptsize}\\
    $\mathit{COE}_{X} = a_{X} + \tilde{\rho}_{X}$
    \begin{scriptsize}\Comment*[f]{density-aware attention}\end{scriptsize}\\
    $\tilde{E}  = \operatorname{ELU}(\mathit{COE}_{X}^{\top }\times \mathit{WX})$
    \begin{scriptsize}\Comment*[f]{feature aggregation}\end{scriptsize}\\
   
    $a_{E} = f_{a_E}(\mathit{WX},\tilde{E},\alpha_{E})$
    \begin{scriptsize}\Comment*[f]{compute attention}\end{scriptsize}\\
    $\tilde{\rho}_{E} = f_{n}(\rho_{E},a_{E})$
     \begin{scriptsize}\Comment*[f]{normalize density}\end{scriptsize}\\
    $\mathit{COE}_{E} = a_{E} + \tilde{\rho}_{E}$  
    \begin{scriptsize}\Comment*[f]{density-aware attention}\end{scriptsize}\\
    $\tilde{X}  = \operatorname{ELU}(\mathit{COE}_{E}^{\top } \times \tilde{E})$ 
    \begin{scriptsize}\Comment*[f]{feature aggregation}\end{scriptsize}\\
    
    \Return $\tilde{X}$\\
\end{algorithm}

   
    
    
   
   

\section{Experiments}

In this section, we conducted extensive experiments to validate the effectiveness of the proposed DA-HGNN on semi-supervised node classification tasks. We first introduce the datasets we choose and the experimental setting. Then we assess the performance of the DA-HGNN, and compare it with the representative graph-based semi-supervised node classification methods. In the ablation study, we validate the effectiveness of density-aware attention weight.

\subsection{Datasets}

To evaluate the effectiveness of the proposed DA-HGNN on semi-supervised node classification tasks, we conduct experiments to test it on three benchmark image datasets,  including MNIST \cite{lecun1998gradient}, CIFAR-10 \cite{krizhevsky2009learning} and SVHN \cite{netzer2011reading}. Each datasets is used in an semi-supervised learning setup where only a small part of data samples are labeled. More details of these datasets and their usages in our experiments are introduced as follow, which also are summarized in \cref{datasets-table}.

\textbf{MNIST.} This dataset contains of 10 classes images of hand-written digits from ‘0’ to ‘9’. It has 60000 training images and 10000 test images, and all the images are grayscale. We randomly select 1000 images for each  digit class and obtain 10000 images at all to conduct our experiments. Similar to the prior work \cite{DBLP:journals/pr/LinKLZC21,DBLP:conf/cvpr/JiangZLTL19}, we use 784-dimension feature vectors converted from  grayscale to represent each digit image.

\textbf{CIFAR-10.} Its training set contains 50000 natural images coming from 10 classes, while its test set involves 10000 images from the same 10 classes. Each image is $32\times32$ with RGB channels. In our experiments, we use the 10000 images from the independent test set to evaluate our method. To represent each image, we use the same 13-layers CNN networks as in \cite{DBLP:conf/cvpr/IscenTAC19, DBLP:conf/nips/TarvainenV17, laine2017temporal} to extract the features.

\textbf{SVHN.} This dataset contains 73257 training and 26032 test images, which can be seen as similar to MNIST. But each image in this dataset is a $32\times32$ RGB image and contains multiple digits. Similar to CIFAR-10 dataset, we randomly select 1000 images for each class from the independent test set to get 10000 images at last for our evaluation. We also use the same 13-layers CNN networks as in \cite{DBLP:conf/cvpr/IscenTAC19, DBLP:conf/nips/TarvainenV17, laine2017temporal} to extract features to represent each image.

\begin{table*}[ht]
\caption{Datasets statistics and the extracted features in experiments.}
\label{datasets-table}
\vskip 0.15in
\begin{center}
\begin{small}
\begin{tabular}{lccccc}
\toprule
Dataset & Training Samples & Validating Samples & Testing Samples & Classes & Features \\
\midrule
MNIST     & 500 / 1000 / 2000 / 3000 / 4000 & 1000 & 8500 / 8000 / 7000 / 6000 / 5000 & 10 & 784 \\
CIFAR-10  & 500 / 1000 / 2000 / 3000 / 4000 & 1000 & 8500 / 8000 / 7000 / 6000 / 5000 & 10 & 128 \\
SVHN      & 500 / 1000 / 2000 / 3000 / 4000 & 1000 & 8500 / 8000 / 7000 / 6000 / 5000 & 10 & 128 \\
\bottomrule
\end{tabular}
\end{small}
\end{center}
\vskip -0.1in
\end{table*}

\begin{table*}[ht]
\caption{Classification accuracy ($\%$) on dataset MNIST, CIFAR-10 and SVHN with different labeled samples.}
\label{comparison-table}
\vskip 0.15in
\begin{center}
\begin{small}
\begin{tabular}{lccccc}
\toprule 
\hline
Datasets & \multicolumn{5}{c}{MNIST} \\ 
No. of labelled Samples & 500 & 1000 & 2000 & 3000 & 4000 \\ 
\hline
GCN    &  90.37±0.32  &  90.42±0.39  &  90.28±0.34  &  90.30±0.38  &  90.20±0.32  \\
GATs   &  91.40±0.14  &  92.44±0.12  &  92.99±0.20  &  93.05±0.18  & 93.41±0.28 \\
HGNN   &  88.70±0.46   &  90.26±0.53  &  91.34±0.45  &  92.25±0.32  & 92.41±0.34 \\
DA-HGNN (ours)        & \textbf{92.09±1.03}   &  \textbf{93.63±0.45}  & \textbf{93.65±0.41}   &  \textbf{94.33±0.43}  &  \textbf{94.34±0.38}  \\
\hline 
\hline
Datasets & \multicolumn{5}{c}{CIFAR-10} \\ 
No. of labelled Samples & 500 & 1000 & 2000 & 3000 & 4000 \\ 
\hline
GCN    & 91.48±0.25  &  91.70±0.12  &  92.55±0.13  &  92.64±0.14  & 92.98±0.16  \\
GATs   & 93.80±0.13  &  93.59±0.42  &  93.90±0.13  &  93.97±0.09  & 93.80±0.04  \\
HGNN   & 90.97±0.41  & 91.26±0.19   &  91.35±0.30  &  91.68±0.11  & 91.85±0.33  \\
DA-HGNN (ours)      &  \textbf{93.84±0.10}   & \textbf{93.98±0.12}   &  \textbf{93.93±0.10} &  \textbf{94.02±0.07}   &  \textbf{93.86±0.09}  \\
\hline 
\hline
Datasets & \multicolumn{5}{c}{SVHN} \\ 
No. of labelled Samples & 500 & 1000 & 2000 & 3000 & 4000 \\ 
\hline
GCN    & 95.52±0.12  &  95.17±0.20  & 95.54±0.11  &  95.70±0.11    & 95.61±0.16  \\
GATs   & 96.15±0.09  &  96.23±0.10  &  96.33±0.04 &   96.38±0.13    &   96.14±0.02  \\
HGNN   & 94.20±0.58 & 94.61±0.53  & 94.73±0.49  & 94.73±0.33  &  94.61±0.35  \\
DA-HGNN (ours)       & \textbf{96.36±0.14}  & \textbf{96.55±0.10}  &  \textbf{96.46±0.22}  &  \textbf{96.46±0.15}  &  \textbf{96.40±0.12} \\
\hline 
\bottomrule
\end{tabular}
\end{small}
\end{center}
\vskip -0.1in
\end{table*}

\subsection{Experimental setting}
 
\textbf{Datasets setting.} For image datasets MNIST, CIFAR-10 and SVHN, we randomly select 50, 100, 200, 300, 400 images per class and get 500, 1000, 2000, 3000, 4000 images as labeled samples. For unlabeled samples, we pick 100 images at random per class and get 1000 images used for validation. And the remaining 8500, 8000, 7000, 6000, 5000 images are used as test samples. Finally, we can obtain the training/validation/testing data splits, including 500/1000/8500, 1000/1000/8000, 2000/1000/7000, 3000/1000/6000 and 4000/1000/5000. 

\textbf{Model setting.} For the model architecture of our DA-HGNN, we employed one hyper-graph convolutional layer and two density-aware hyper-graph attention layers, where the first density-aware hyper-graph attention layer is a multi-head attention with 4 heads. The number of units in hyper-graph convolutional hidden layer is set to 256 for MNIST dataset and 64 for CIFAR-10 and SVHN datasets. The number of units in the density-aware hyper-graph attention hidden layer is set to 8. The value of the predefined threshold $\delta$ is set to 0.4. We employ the Xavier algorithm \cite{DBLP:journals/jmlr/GlorotB10} for the initialization of $W$, $\alpha_{X}$ and $\alpha_{E}$.
We adopt the Adam optimizer \cite{DBLP:journals/corr/KingmaB14} with learning rate 0.002, and the learning rate decays to half after every 100 epochs. We train DA-HGNN for a maximum of 3000 epochs, and stop training if the validation loss does not decrease for 100 consecutive epochs, as suggested in work \cite{DBLP:conf/iclr/KipfW17, DBLP:conf/cvpr/JiangZLTL19}.

\subsection{Performance on Semi-supervised Node Classification}

\textbf{Baselines.} We compare the  proposed DA-HGNN with representative graph-based semi-supervised node classification methods, including GCNs \cite{DBLP:conf/iclr/KipfW17}, GATs \cite{DBLP:conf/iclr/VelickovicCCRLB18}, HGNN \cite{DBLP:conf/aaai/FengYZJG19}. For a fair comparison, we construct a k-nearest neighbor graph for all the methods, and the value of $k$ is set to 10. We retrain all the baseline method, and all the reported result are averaged over 10 runs.

\begin{figure}[htbp]
\vskip 0.2in
\begin{center}
\centering
\subfigure[GATs result]{
\begin{minipage}[t]{0.5\linewidth}
\centering
\includegraphics[width=1.6in]{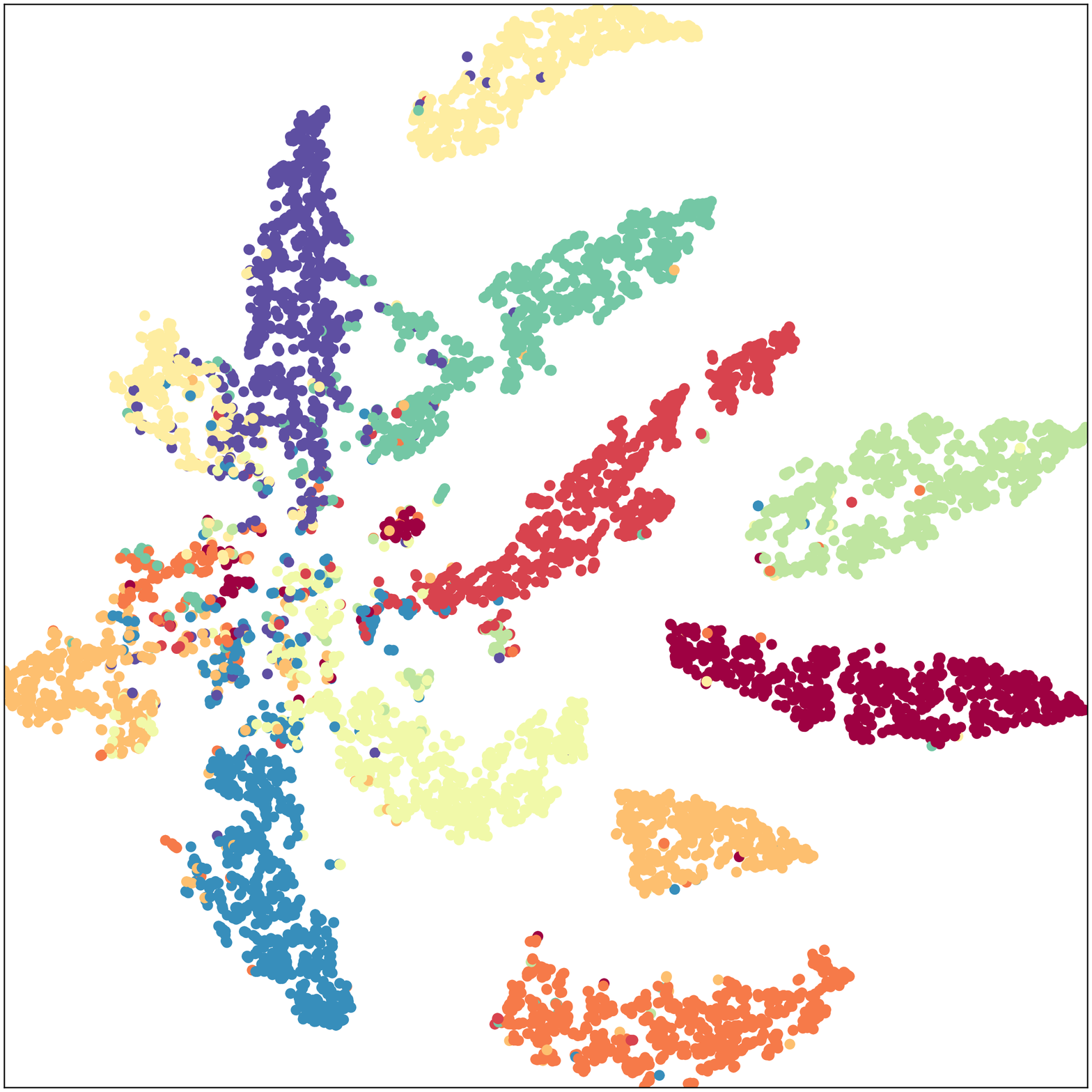}
\end{minipage}%
}%
\subfigure[DA-HGNN result]{
\begin{minipage}[t]{0.5\linewidth}
\centering
\includegraphics[width=1.6in]{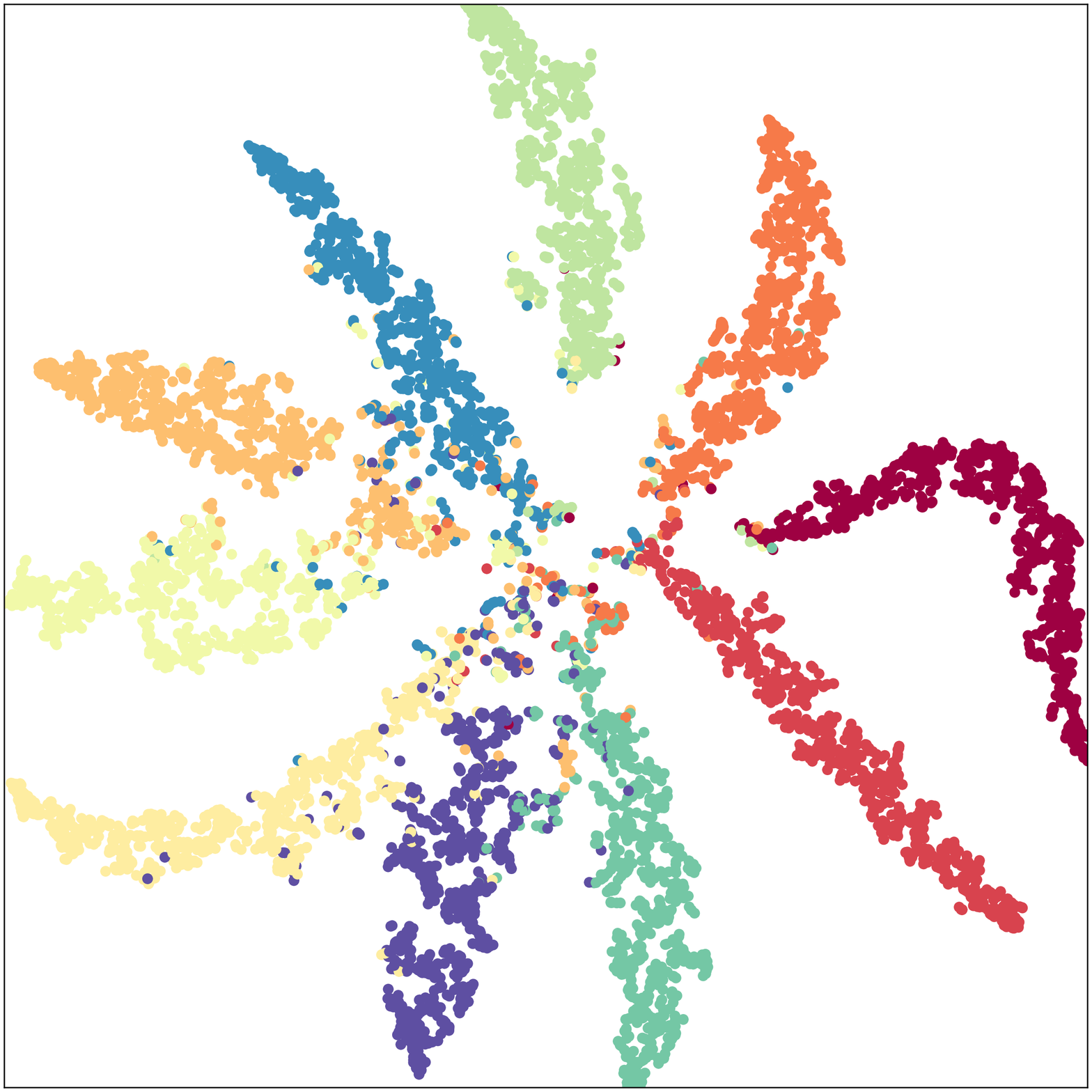}
\end{minipage}%
}%
\caption{2D t-SNE \cite{van2008visualizing} visualizations of the feature embedding of GATs and the proposed DA-HGNN respectively on MNIST dataset with 1000 labeled samples. Each dot in the figure corresponds to one sample, and different classes are marked by different colors.}
\label{t-SNE}
\end{center}
\vskip -0.2in
\end{figure}

\textbf{Results.} \cref{comparison-table} summarizes the classification accuracy comparison results on three image datasets widely used in many visual classification tasks. The best results are marked. From these results, we can observe that: (1) DA-HGNN outperforms the baseline method GCNs by significant margins on all datasets. This demonstrates that, by modeling the correlation among data as a hyper-graph and performing density-aware attention neighborhood aggregation, DA-HGNN is more accurate on semi-supervised node classification; (2) DA-HGNN significantly outperforms HGNN on all image datasets, which straightforwardly indicates the higher predictive accuracy on semi-supervised node classification of DA-HGNN by performing density-aware attention feature aggregation; and (3) DA-HGNN performs better than GATs, which further demonstrates the effectiveness of density-aware attention neighborhood aggregation. To have a more intuitive comparison, we further visualize the learned feature embeddings of GATs and the proposed DA-HGNN, respectively, on MNIST dataset with 1000 labeled samples, which is shown in \cref{t-SNE}. We can observe that, the data of different classes are distributed more clearly and compactly in our DA-HGNN feature representation learning.

\begin{table*}[ht!]
\begin{center}
\caption{Effectiveness of density-aware attention weight on dataset MNIST.}
\label{ablation-study-table}
\vskip 0.15in
\begin{small}
\begin{tabular}{lccccc}
\toprule 
Datasets & \multicolumn{5}{c}{MNIST} \\ 
No. of labelled Samples & 500 & 1000 & 2000 & 3000 & 4000 \\ 
\hline
DA-HGNN w/o density\tnote{*}    &  91.11±1.36  &  92.57±0.86 &  93.63±0.50 & 94.14±0.45  &  94.21±0.38 \\
DA-HGNN with density\tnote{*}       & \textbf{92.09±1.03}   &  \textbf{93.63±0.45}  & \textbf{93.65±0.41}   &  \textbf{94.33±0.43}  &  \textbf{94.34±0.38}  \\
\bottomrule
\end{tabular}
\end{small}
\vskip -0.1in
\end{center}
\end{table*}

\subsection{Ablation Study}

We propose to integrate the density information of each node and each hyperedge as a part of attention weight. To evaluate the effectiveness of the density-aware attention weight, we conduct an ablation study. We remove the density information from the proposed DA-HGNN and only keep the traditional attention mechanism which only consider feature similarity, and we mask it as DA-HGNN w/o density. That means, the DA-HGNN w/o density only has one hyper-graph convolutional layer and two traditional hyper-graph attention layers. The proposed version is labeled as DA-HGNN with density. We conduct ablation experiments on MNIST dataset, and the comparison results between DA-HGNN w/o density and DA-HGNN with density are shown in \cref{ablation-study-table}. All the reported results are averaged over 10 runs. From these results, we can note that integrating the density information of each node and hyperedge as a part of attention weight can significantly improve the performance of hyper-graph attention neighborhood aggregation and achieve higher predictive accuracy on semi-supervised node classification.

\section{Conclusion}

In this paper, we propose a novel Density-Aware Hyper-Graph Neural Networks (DA-HGNN) for semi-supervised node classification. To better explore the high-order semantic connection relation among data in semi-supervised learning, we propose the use of hyper-graph to model the correlation. We define a density rule for each node and hyperedge. Then we learn the density information and integrate it as a part of attention weight via a simple and effective way. By combining the density information, we can exploit it explicitly for hyper-graph representation learning and achieve more accurate attention neighborhood aggregation. We have conducted extensive experiments on three image datasets and demonstrated the validness of density-aware attention weight and the effectiveness of the proposed DA-HGNN on various semi-supervised node classification tasks.



\bibliography{DAHGNN_paper}
\bibliographystyle{icml2022}

\end{document}